\documentclass[letterpaper, 10 pt, conference]{ieeeconf}
\IEEEoverridecommandlockouts                              
\usepackage{times}
\usepackage{epsfig}
\usepackage{graphicx}
\usepackage{amsmath}
\usepackage{amssymb}

\usepackage{graphicx}
\usepackage{epsfig}
\usepackage{epic,eepic}
\usepackage{times}
\usepackage{mathptmx}
\usepackage{amsfonts}
\usepackage{amsmath}
\usepackage{cite}
\usepackage{color}

\usepackage{times}
\usepackage{multicol}

\definecolor{red}{rgb}{1,0,0}
\definecolor{green}{rgb}{0,1,0}
\definecolor{blue}{rgb}{0,0,1}
\definecolor{violet}{rgb}{1,0,1}
\definecolor{cyan}{cmyk}{1,0,0,0}
\definecolor{magenta}{cmyk}{0,1,0,0}
\definecolor{yellow}{cmyk}{0,0,1,0}

\definecolor{white}{rgb}{1,1,1}

\usepackage{jumoline}

\newcommand{\CO}[1]{}

\newcommand{\CommentOut}[1]{}

 \newcommand{\editage}[1]{}

\onecolumn

\begin{document}

\newcommand{\FIG}[3]{
\begin{minipage}[b]{#1cm}
\begin{center}
\includegraphics[width=#1cm]{#2}\\
{\scriptsize #3}
\end{center}
\end{minipage}
}

\newcommand{\FIGU}[3]{
\begin{minipage}[b]{#1cm}
\begin{center}
\includegraphics[width=#1cm,angle=180]{#2}\\
{\scriptsize #3}
\end{center}
\end{minipage}
}

\newcommand{\FIGm}[3]{
\begin{minipage}[b]{#1cm}
\begin{center}
\includegraphics[width=#1cm]{#2}\\
{\scriptsize #3}
\end{center}
\end{minipage}
}

\newcommand{\FIGR}[3]{
\begin{minipage}[b]{#1cm}
\begin{center}
\includegraphics[angle=-90,width=#1cm]{#2}
\\
{\scriptsize #3}
\vspace*{1mm}
\end{center}
\end{minipage}
}

\newcommand{\FIGRpng}[5]{
\begin{minipage}[b]{#1cm}
\begin{center}
\includegraphics[bb=0 0 #4 #5, angle=-90,clip,width=#1cm]{#2}\vspace*{1mm}
\\
{\scriptsize #3}
\vspace*{1mm}
\end{center}
\end{minipage}
}

\newcommand{\FIGpng}[5]{
\begin{minipage}[b]{#1cm}
\begin{center}
\includegraphics[bb=0 0 #4 #5, clip, width=#1cm]{#2}\vspace*{-1mm}\\
{\scriptsize #3}
\vspace*{1mm}
\end{center}
\end{minipage}
}

\newcommand{\FIGtpng}[5]{
\begin{minipage}[t]{#1cm}
\begin{center}
\includegraphics[bb=0 0 #4 #5, clip,width=#1cm]{#2}\vspace*{1mm}
\\
{\scriptsize #3}
\vspace*{1mm}
\end{center}
\end{minipage}
}

\newcommand{\FIGRt}[3]{
\begin{minipage}[t]{#1cm}
\begin{center}
\includegraphics[angle=-90,clip,width=#1cm]{#2}\vspace*{1mm}
\\
{\scriptsize #3}
\vspace*{1mm}
\end{center}
\end{minipage}
}

\newcommand{\FIGRm}[3]{
\begin{minipage}[b]{#1cm}
\begin{center}
\includegraphics[angle=-90,clip,width=#1cm]{#2}\vspace*{0mm}
\\
{\scriptsize #3}
\vspace*{1mm}
\end{center}
\end{minipage}
}

\newcommand{\FIGC}[5]{
\begin{minipage}[b]{#1cm}
\begin{center}
\includegraphics[width=#2cm,height=#3cm]{#4}~$\Longrightarrow$\vspace*{0mm}
\\
{\scriptsize #5}
\vspace*{8mm}
\end{center}
\end{minipage}
}

\newcommand{\FIGf}[3]{
\begin{minipage}[b]{#1cm}
\begin{center}
\fbox{\includegraphics[width=#1cm]{#2}}\vspace*{0.5mm}\\
{\scriptsize #3}
\end{center}
\end{minipage}
}



\newcommand{\acprPaperID}{25}





\author{Sugimoto Takuma ~~~ Yamaguchi Kousuke ~~~~ Tanaka Kanji 
\thanks{Our work has been supported in part by JSPS KAKENHI Grant-in-Aid for Scientific Research (C) 26330297,
and (C) 17K00361.}
\thanks{The authors are with Graduate School of Engineering, University of Fukui.
{\tt\small tnkknj@u-fukui.ac.jp}}%
\vspace*{-5mm}}

\newcommand{\vs}{\hspace*{-1mm}}

\renewcommand{\FIGpng}[5]{
\begin{minipage}[b]{#1cm}
\begin{center}
\includegraphics[bb=0 0 #4 #5, clip, width=#1cm]{kd19_yamaguchi/#2}\vspace*{-1mm}\\
{\scriptsize #3}
\vspace*{1mm}
\end{center}
\end{minipage}
}

\newcommand{\figA}[1]{
\begin{figure}[t]
\begin{center}
\FIG{15}{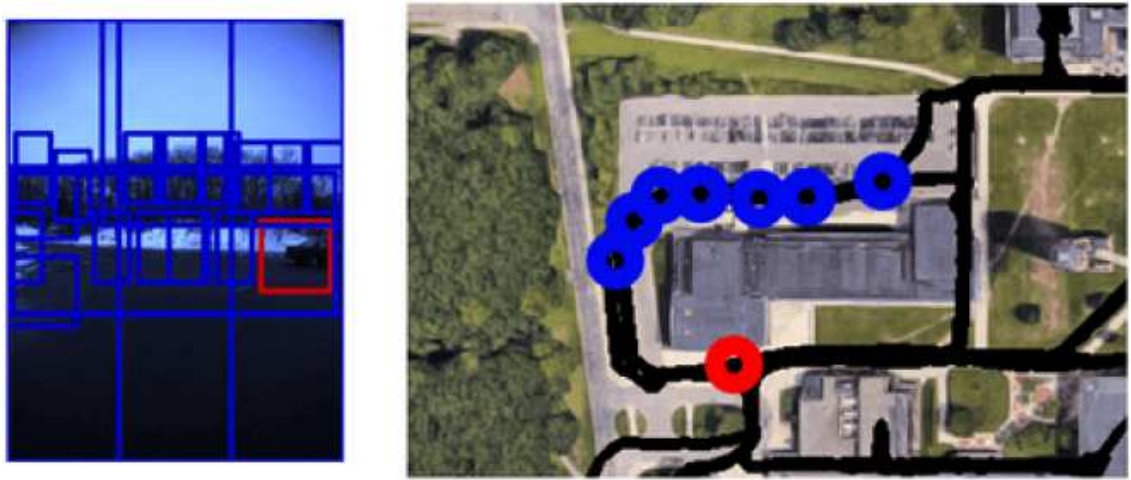}{}
\caption{We detect significant changes as inconsistency between strong and weak self-localization modules in visual SLAM. In the figure, 
blue and red 
colors indicate self-localization results 
for the strong and weak BoW features.}\label{fig:A}
\end{center}
\vspace*{-7mm}
\end{figure}
}

\newcommand{\FIGSpng}[5]{
\begin{minipage}[b]{#1cm}
\begin{center}
\includegraphics[bb=0 0 #4 #5, clip, height=#1cm, width=#1cm]{#2}\vspace*{-1mm}\\
{\scriptsize #3}
\vspace*{1mm}
\end{center}
\end{minipage}
}

\newcommand{\forfigC}[2]{
 \begin{minipage}{0.6cm}
   \FIGpng{0.6}{ronbun/yamaguchi/#1.png}{#2}{486}{723}
 \end{minipage}
}

\newcommand{\figC}[1]{
\begin{figure}[t]
 \begin{center}
   \FIG{15}{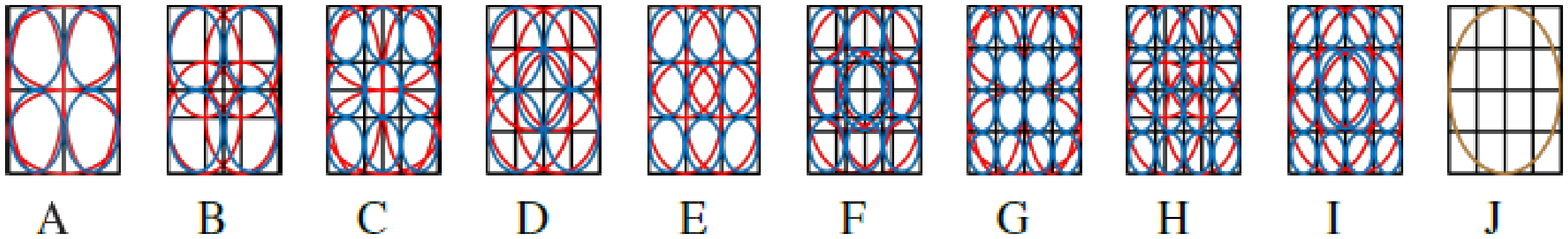}{}
 \caption{Library of OBBs. A grid with cells (black) is imposed on the image region and OBBs with 
different combinations of the cells (red/blue circles) are defined.}\label{fig:C}
 \end{center}
\end{figure}
}

\newcommand{\tabD}[1]{
\begin{table}[t]
  \begin{center}
  \caption{Change detection results.}\label{tab:D}
  \begin{tabular}{r|rrrr|}
 $X$ [\%]      	&\vs 5	&\vs 10	&\vs 15	&\vs 20 \\ \hline 
     
    FD	&\vs 7.8	&\vs 15.8	&\vs 27.2	&\vs 39.6	\\
    PC	&\vs 0.0	&\vs 2.1	&\vs 7.2	&\vs 19.6	\\ 
    AD	&\vs 0.4	&\vs 2.5	&\vs 10.6	&\vs 25.2	\\
    FD+AD+PC &\vs 2.6 &\vs 20.1 &\vs 34.1 &\vs 42.1	\\
    FD+AD &\vs 2.6 &\vs 19.7 &\vs 33.8 &\vs 41.9 \\
    FD+PC &\vs 3.1 &\vs 11.0 &\vs 20.6 &\vs 34.9 \\ \hline
  \end{tabular}
  \end{center}
\vspace*{-5mm}
\end{table}
}

\newcommand{\figD}[1]{
\begin{figure}
  \begin{center}
\FIG{15}{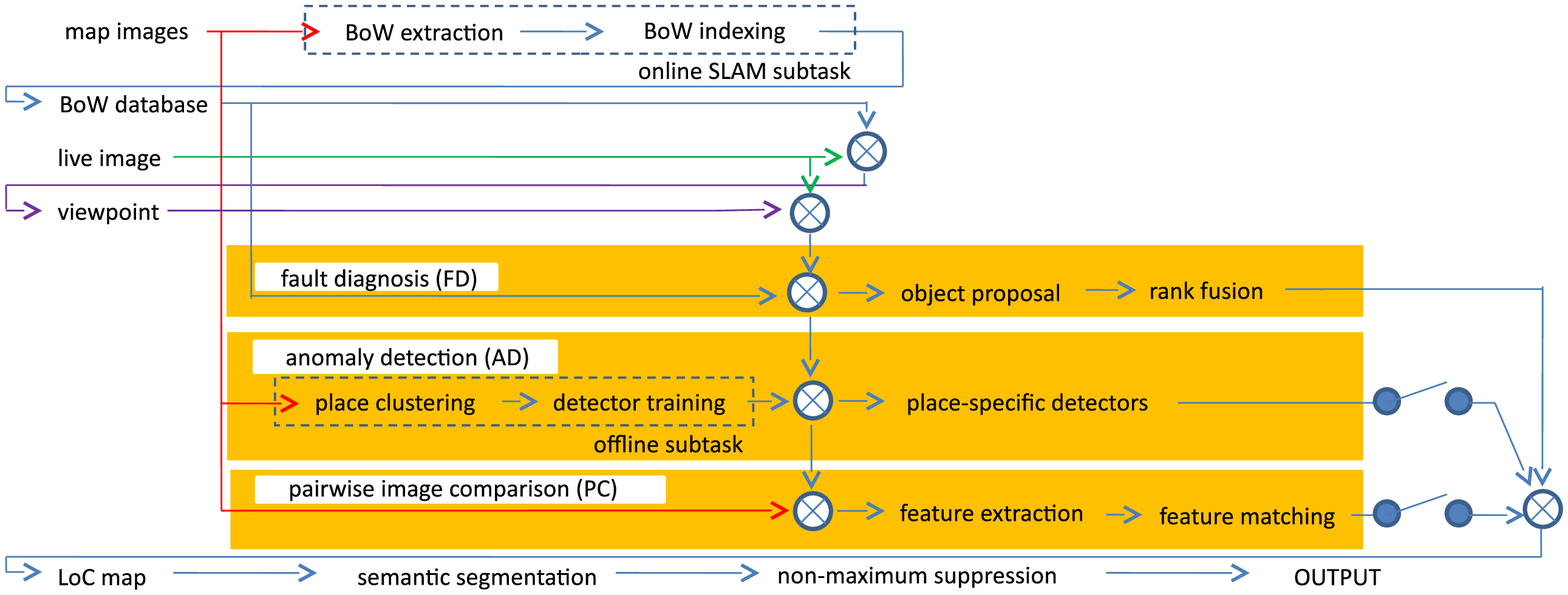}{}
    \caption{System architecture.}\label{fig:D}
  \end{center}
\vspace*{-5mm}
\end{figure}
}

\newcommand{\forfigE}[1]{
 \begin{minipage}{1.3cm}
  \begin{center}
\FIGpng{1.3}{ronbun/fig2_fix/#1}{}{170}{223}
  \end{center}
 \end{minipage}
}

\newcommand{\figE}[1]{
\begin{figure}[t]
 \begin{center}
\FIG{15}{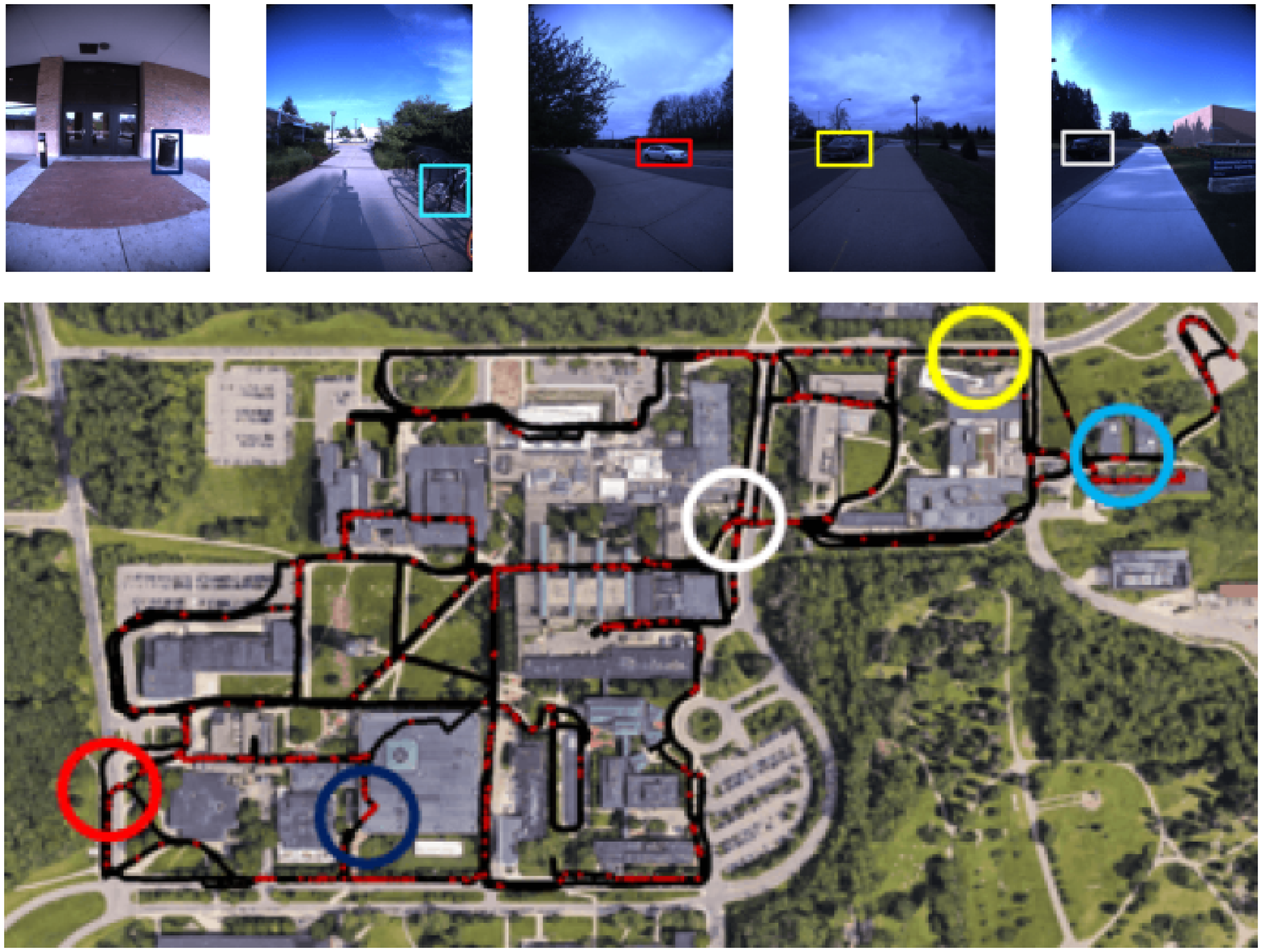}{}
 \caption{Experimental environments and robot trajectories.
Colored rectangles and circles 
respectively
indicate 
the bounding boxes 
and 
the GPS locations
of the ground-truth change objects.
Different colors
represent
different change objects.} \label{fig:E}
 \end{center}
\vspace*{-5mm}
\end{figure}
}

\title{\LARGE \bf
Fault-Diagnosing SLAM for Varying Scale Change Detection
}

\maketitle

\begin{abstract}
In this paper, we present a new fault diagnosis (FD) -based approach 
for detection of imagery changes 
that can detect significant changes as inconsistencies between different sub-modules (e.g., self-localizaiton) of visual SLAM.
Unlike classical change detection approaches 
such as pairwise image comparison (PC) and anomaly detection (AD), 
neither the memorization of each map
image nor the maintenance of up-to-date place-specific anomaly detectors are required in this FD approach.
A significant challenge that is encountered
when incorporating different SLAM sub-modules into FD involves dealing with the varying scales of objects that have changed (e.g., the
appearance of small dangerous obstacles on the floor). To address this issue, we reconsider the bag-of-words (BoW) image representation, by
exploiting its recent advances in terms of self-localization and change detection. As a key advantage, BoW image representation can be
reorganized into any different scaling by simply cropping the original BoW image. Furthermore, we propose to combine different self-localization
modules with strong and weak BoW features with different discriminability, and to treat inconsistency between strong and weak
self-localization as an indicator of change. The efficacy of the proposed approach for FD with/without AD and/or PC was experimentally
validated.
\end{abstract}

\section{Introduction}\label{sec:intro}

For long-term map maintenance in dynamic environments,
a robotic visual SLAM system must detect changed objects (e.g., furniture movement, 
and building construction) in a live image with respect to the map,
while ignoring nuisance changes (e.g., sensor noises, registration errors, 
and occlusions) 
during long-term multi-session navigation.
One approach is to formulate the problem as a
pair-wise image comparison (PC), to compare each live-map-image-pair using image differencing techniques \cite{accv12}.
However, this requires that a robot memorizes every map image; hence, scaling to large-size environments is difficult. An alternative
approach is to formulate the problem as anomaly detection (AD), and predict anomalies (changes) in a live image with
respect to a pre-trained normal model (map) \cite{hinton2006reducing}.
However, this requires a robot to re-train the normal model frequently
to keep it up-to-date every time the map is updated.

An alternative is to formulate the problem as a fault diagnosis (FD) to treat inconsistency (changes) between the responses of different sub-modules of SLAM (e.g., map-relative self-localization vs. pose-tracking \cite{JacobsonCM15}, self-localization vs. mapping \cite{iv18mapmaintenance})
as an indicator of the likelihood-of-changes (LoC). 
The first solution to a multi-experience based 
mapping system was pioneered by \cite{churchill2013experience}, 
which maintains a collection of mapping sub-modules using data from differing
environmental conditions (i.e., visual experiences).
If self-localization with respect to such a map performs sufficiently well, it satisfies the encountered conditions and there is no significant change. However, if localization performance is poor, the map does not satisfy the conditions and environmental changes can occur with high probability. Such a new FD-based change detection framework has two main advantages:
\begin{itemize}
\item
No additional storage or detector engine (but only existing map database and localization engine) is required; 
\item
Degradation of map quality (i.e., need for map update) in terms of localization performance can be measured.
\end{itemize}

\figA

A significant challenge when incorporating different SLAM sub-modules into FD involves addressing varying scales
of changed objects. This is an important issue because a robot is often required to inspect not only image-level but also
sub-image-level changes, such as the appearance of small dangerous obstacles on the floor \cite{nava2019learning}. 
Typical feature extractors used by self-localization
(e.g., ConvNet \cite{SunderhaufSDUM15}, autoencoders \cite{MerrillH18})
utilize live images as the input at the same scaling as the map images. 
If
they are tested with varying scale images, they tend to fail, even when there is no change. As such, it is difficult to apply them
to sub-image-level change detection.

In this paper,
we reconsider the bag-of-words (BoW) image representation based on recent advances in self-localization
techniques (Fig. \ref{fig:A}). We consider that BoW has several desirable properties. Firstly, a recently developed BoW-based method
has developed into the state-of-the-art in self-localization \cite{Garcia-Fidalgo2018}. 
Secondly, BoW is a compressed (and yet discriminability) image
representation that assists in the suppression of the map maintenance cost. Thirdly, BoW has a good affinity to codebook-based
image representation \cite{kim2005real},
which has attracted increasing attention in the field of change detection \cite{shah2015self}.
Most
importantly, the full-image-level BoW representation can be flexibly reorganized into sub-image-level BoW representation
by simply cropping the BoW image with a smaller ROI.

In this contribution, we present a new single-view FD-based change detection approach that can detect sub-image-level
changes while simultaneously localizing the robot-self. 
The idea is to crop the original BoW image with different
ROIs to reorganize it into different sub-image-level BoW features. Specifically, we propose two different types of
sub-image-level BoW features: strong and weak, with different levels of discriminability. The former
are discriminability features with large ROIs that are useful for reliable self-localization.
It should be noted that such strong
features on their own are
{\it not} 
sufficient for the FD-based change detection because strong features merely cause a self-localization
fault, which is required by the FD-based change detection. Therefore, we introduce weak features with small object-level
ROIs to trigger a self-localization fault intentionally. Intuitively, changes are expected to be detected as inconsistencies between
such strong and weak self-localization modules.
Experiments on challenging cross-season change detection using publicly
available NCLT dataset
\cite{nclt}
validate the efficacy of the proposed approach of FD with/without combining AD and/or PC.

\section{Approach}

The main focus of this paper is to introduce a novel fault-diagnosis -based approach to improve
object-level change detection. As mentioned in the 
Section \ref{sec:intro}, existing techniques can be categroized into three groups,
PC (\ref{sec:pc}), AD (\ref{sec:ad}), and FD (\ref{sec:fd}).
It should be noted that
PC and AD are not always available,
depending on 
the availablity of map images 
and
the update frequency of place-specific anomaly detectors.
Therefore,
there are 
four possible combinations
of 
available change detection
modules,
including
FD, FD+AD, FD+PC, and FD+AD+PC (Fig. \ref{fig:D}).
Each of these modules
will be investigated in the experimental section.
It should be noted 
that 
in all change detection approaches, 
knowledge of the current viewpoint is required, 
to pair a live query image with the appropriate map image or the place-specific models.
To this end,
a robust 
viewpoint localization
scheme is also introduced in \ref{sec:fd}.

\subsection{Pairwise Image Comparision (PC)}\label{sec:pc}

For PC,
we introduce a SIFT feature-based PC approach.
Based on the literature (e.g., \cite{accv12}), 
the LoC of a query live feature
is measured
according to
its dissimilarity
to the most similar normal feature.
Firstly,
every live/map image
is represented as
a collection of SIFT features with Harris-Laplace keypoints \cite{SiftOrg}.
The LoC at each keypoint in the query image
is measured using
the L2 distance 
between a SIFT descriptor at that keypoint and its 
nearest-neighbor map SIFT in the 128-dim SIFT feature space.

\subsection{Anomaly Detection (AD)}\label{sec:ad}

In contrast, the AD approach formulates the problem as 
a one-class classification \cite{OCSVM}, 
in which the goal is to classify 
(not a live-map-image-pair but) a query live image as a ``change" 
or a ``no-change" \cite{kim2005real,shah2015self,wu2010spatio,ccr5, palazzolo2018icra,8268055,icra18burgard8,icra18burgard17,babaee2018deep,remotesensing,christiansen2016deepanomaly}, with respect to an offline pretrained normal model. 
Unlike PC approaches,
this 
formulation 
facilitates
the utilization of 
compressed normal models such as 
bag-of-visual-features (BoVFs) \cite{kim2005real,shah2015self,wu2010spatio}, 3D/landmark/grid maps \cite{icra18burgard8,icra18burgard17,ccr5, palazzolo2018icra, 8268055}, compact manifold learning \cite{babaee2018deep,remotesensing,christiansen2016deepanomaly}, 
and autoencoders (AEs) \cite{hinton2006reducing}.
Early studies employed
one-class support vector machines \cite{OCSVM},
or support vector data description \cite{SVDD}.
However, 
their computational scaling 
was poor because of 
the construction and manipulation of the kernel matrix.
Subspace-based methods
\cite{aggarwal2001outlier}, 
\cite{zhang2004hos},
\cite{assent2008inscy},
\cite{nguyen2011unbiased}, 
\cite{kriegel2009outlier}, 
\cite{muller2011statistical}, 
\cite{keller2012hics}
are effective means of finding
anomalous objects in relevant subspaces that are not anomalous in the full-dimensional space.
However, most existing methods use shallow methods
that
typically require substantial feature engineering.
Recently,
deep AEs \cite{hinton2006reducing}
have developed into a predominant approach for learning-based AD.
In this context,
AEs are used differently in two approaches:
(1) mixed
approaches \cite{DBLP:journals/corr/SabokrouFFK16},
in which the learned embeddings are plugged into classical AD methods,
and
(2) fully deep
approaches, 
in which 
the reconstruction error (RE) is directly utilized as an anomaly score \cite{DBLP:journals/corr/SabokrouFFK16}. 
Our approach belongs to the latter \cite{DBLP:journals/corr/SabokrouFFK16, SAE, VAE, makhzani2015winner}.

The basic idea is
to reconstruct a query live image $I$
by using its counterpart (or linked) normal model (i.e., AE) $c_{j}$.
The AE is designed to extract
the common factors of variation from normal samples
and to reconstruct them accurately.
As such,
anomalous samples
do not contain these common factors of variation; 
hence, they 
cannot be 
accurately reconstructed.
Therefore, 
the pixel-level LoC 
for a given image region $P$ 
can be evaluated
by the RE at each pixel:
$V_{RE}(P) = \sum_{p\in P} | I(p) - I'(p) | $, 
in which
the images $I$ and $I'$
are the input image and the image 
is reconstructed by 
the AE
that is linked to the corresponding map image,
respectively,
and 
$|\cdot |$
is an absolute value operator.
If
$V_{RE}$
exceeds a pre-defined threshold $V_{RE}^*$,
the region of interest
$P$
is determined to be an anomalous object.

A place-specific
anomaly detector
is required
to define a-priori 
what constitutes a place.
Recently,
we have explored several 
place definition approaches
in AE-based change detection \cite{itsc19yamaguchi}.
In the current study,
we employ the k-means clustering approach
to partition the map image set into $k$ place-classes.
Each of the $k$ trained AEs 
is used as the 
normal model
of the map images that belong to the cluster.

\figD

A notable design issue in training place-specific AEs
is the determination of 
the
threshold $V_{RE}^*$
on $V_{RE}$
for different place-specific AEs.
The RE outputs by
different AEs
are not comparable
because
individual AEs are trained using
different training sets.
To address this issue,
we normalize
each RE value
by the AE-specific normalizer constant.
In the normalization process,
we approximate
the probability distribution function (PDF)
of the REs
using a Gaussian distribution,
and normalize the RE value by
subtracting the mean value $\mu$
and dividing by 
its standard deviation (SD) $\sigma$
and by a normalizer coefficient $c$
with a value of $c=0.8$ as the default.
This normalization
allows
outputs from different AEs
to be directly compared.

\subsection{Fault Diagnosis (FD)}\label{sec:fd}

The basic idea of the FD-based approach
is to introduce
strong and weak self-localization modules
based on different levels of feature discriminativity.
Inconsistencies between
responses from the strong and weak self-localization
are then treated
as a change indicator.
Formally,
such a difference in discriminability
can be realized by varying 
the number of 
used BoW sub-images 
between
the strong and weak self-localization modules.
Detailed explanations
of sub-image extraction, 
weak self-localization,
strong self-localization,
fault diagnosis
and 
change detection
is presented in the following.

For sub-image extraction,
we employ
both supervised and unsupervised
object proposal approaches
that are expected to act as strong and weak features, respectively.
The supervised YOLO method from \cite{yolo}
is employed
to extract 
1-11 OBBs per image (Fig. \ref{fig:A}).
The unsupervised 
BING
proposal method from \cite{cheng2014bing}
is employed 
to extract
42-50
class-agnostic
OBBs per image (Fig. \ref{fig:A}).
In addition,
we introduce 
a various combinations 
of non-adaptive fixed OBBs
as shown in Fig. \ref{fig:C}.
A natural 
design choice
is to use 
the weak and strong features
only for
the weak and strong self-localization modules,
respectively.
However,
we propose to use every feature-type for 
both self-localization modules,
which
works better in practice.

For weak self-localization,
we 
use a single sub-image-level BoW 
(without enhancing it using the available contextual information)
to trigger a self-localization fault intentionally,
when and only when there is a significant change in the sub-image region.
Specifically,
we use
the recently developed state-of-the-art BoW framework 
in \cite{Garcia-Fidalgo2018}.
In a previous work,
we 
studied this framework in 
a different context of simultaneous mapping and localization (i.e., SLAM) \cite{YamamotoTT19}.
In the current study,
we extend this framework to sub-image-level BoW
and implement 
mapping and localization 
as two separate (i.e., offline and online) proccesses 
rather than a single SLAM process.
Formally,
in
the offline map-building process,
a collection of ORB features 
\cite{mur2015orb}
is extracted from each map sub-image
and then indexed to the inverted file.
Simultaneously,
we update
the BoW vocabulary 
by incrementally incorporating newly arrived visual words.
In the online localization process,
the inverted file 
is retrieved 
using each word in each live sub-image as a query,
and 
each retrieval result
is further refined by 
the TF-IDF scoring scheme \cite{sivic2003video},
the ratio test \cite{lowe2004distinctive},
RANSAC geometric verification, 
and island clustering \cite{dloopdetector}.

For strong self-localization,
we 
extend
the weak self-localization framework
to aggregate 
multiple BoW sub-images 
to increase robustness and discriminability.
It should be noted
that a weak self-localization module
outputs
a ranked list of 
all the map images
in the order of relevance.
For strong self-localiation,
these weak self-localization modules are ensembled
and the ranked lists are aggregated into a single strong rank list
using the robust 
rank aggregation technique in \cite{kanji2015unsupervised}.
As such,
the rank values $r_1$, $\cdots$, $r_N$
of a given map image
from $N$ different self-localization modules 
are aggregated
into
a relevance score
by rank fusion:
$\sum_{i=1}^N r_i^{-1}$.

For FD,
we evaluate
inconsitency between 
the different ranked lists 
output 
using
weak and strong self-localization modules.
Given a relevant map-image top-ranked 
by strong self-localization,
inconsitency can be evaluated 
by the rank of that relevant map image in the weak rank list.
As such,
a larger rank value 
represents
greater inconsistency.
To address the inherent retrieval noise in the strong self-localization,
we consider multiple top-$Y$ ranked map images ($Y=10$) in the strong rank list,
and then the corresponding $Y$ rank values in the weak rank list 
are aggregated into the final decision.
For the aggregation,
we propose to use pixel-wise min pooling.
This is performed
because 
the retrieval noise 
influences
and can 
spuriously increase the rank values.
This increases
the difficulty
associated with
implementing
the other typical pooling techniques
such as max or average pooling.

\figC

For change detection,
we aggregate
the sub-image-level rank values 
into image-level LoC maps
by incorporating individual OBBs.
The rank aggregation 
problem
was explored in the
context of
part-based self-localization
in our previous study \cite{kanji2015unsupervised}.
The method used in the current study 
is based on
our previous method
with a few key modifications:
Firstly, our previous study 
emphasized at image-level ranking,
whereas this study aims to obtain pixel-level rank values.
Secondly, the previous method 
utilized
non-overlapping query live sub-images (from color-based segmentation)
as inputs,
whereas the current
method utilizes
overlapping query sub-images (unsupervised/supervised OBBs)
with {\it variable}
amounts of overlap per pixel.
To address this issue,
we must perform a new task of pixel-wise rank fusion.
Formally,
we adopt
the recently presented extension of 
{\it variable}
length rank lists
for multi-media retrieval \cite{mourao2015multimodal},
and fuse per-pixel ranking results as follows \cite{kanji2019detection}:
\begin{equation}
r[p]=|J[p]| \left(\sum_{j\in J[p]} r_{j}[p]^{-1} \right)^{-1}.
\end{equation}
$J[p]$
is the set of identifiers of OBBs 
to which pixel $p$ belongs to.
$r_j$ is a rank value of the $j$-th OBB in the map image.

\section{Experiments}\label{sec:exp}

We
evaluated
the proposed change detection framework
for
a challenging
cross-season
scenario.

\subsection{Settings}\label{sec:settings}

We used
a large-scale long-term autonomy dataset,
North Campus long-term (NCLT) dataset,
which is publicly available in \cite{nclt}. 
The data 
used
in this study includes view image sequences 
along 
vehicle trajectories acquired 
using the front facing camera (i.e., Ladybug3) of a Segway vehicle platform (Fig. \ref{fig:E}).
The image size is 1232$\times$1616.
This dataset includes
various types
of changing images
such as cars, pedestrians, building construction, construction machines, posters, tables and whiteboards with wheels,
from seamless indoor and outdoor navigations.
Additionally,
it
has recently
gained significant popularity
as a benchmark in
the SLAM community  \cite{jmangelson-2018a}.

In this study, we used four datasets
labeled
``2012/1/22 (WI)", ``2012/3/31 (SP)", ``2012/8/4 (SU)", and ``2012/11/17 (AU)"
that were collected
from four different seasons.
For all the possible 12 live-map-season-pairs,
we annotated 986 different changing objects
with bounding boxes.
Additionally,
we prepared a collection of 1,973 random destructor images
that
were independent
of the 986 annotated images.
These images did not include changing objects.
We then merged
the 1,973 destructor images
and 986 annotated images to obtain
a map database containing
2,959 images.
Fig. \ref{fig:E}
presents
examples of changing objects
in the dataset.

\subsection{Performance Results}

The performance for
the change detection task
was evaluated in terms of
top-$X,Y$ accuracy.
Firstly,
we estimated
an LoC image
using a change detection algorithm
on the top-$Y$
self-localization hypotheses ($Y=10$).
We then imposed a 2D grid with $10\times 10$ pixel sized cells
on the query image
and estimate an LoC for each cell by max-pooling the pixel-wise LoC values from
all pixels 
that belong to that cell.
To fuse 
results from multiple change detection algorithms (e.g., FD+AD+PC),
the rank fusion algorithm 
in \ref{sec:fd}
is reused.
Next, all the cells from all the map images are sorted
in descending order of LoC,
and the
accuracies of the top-$X$ items in the list were evaluated.
For a specific $X$ threshold,
a successful detection is defined as a changed object 
with an annotated bounding box 
that is sufficiently covered 
(intersection-over-union $\ge$ 50\%) by the top-$X$ percent of 
the cells.

For OBBs for self-localization,
the combination of
J+B+BING+YOLO
outperforms
the other setting.
This combination is used 
for viewpoint localization
in the following change detection experiments.

The number of place-specific AEs
is set as $k=10$
based on the preliminary experiments in \cite{itsc19yamaguchi}.

For performance evaluation,
we further introduce post-processing on the LoC map,
which
can stably boost
the proposed and all the change detection comparison techniques.
The idea is to
use 
recent 
deep learning based
semantic segmenation 
to
detect non-interesting regions with ``sky" or ``ground" labels,
where no interesting change is expected.
LoC values for these non-interesting regions are reset to 0.
Formally,
we employed
a DeepLabV3+ model in \cite{ChenZPSA18}
that 
combines the advantages
of spatial pyramid pooling and encoder-decoder structure.
Input images are resized to 512$\times$512,
and the model was trained on the Cityscaples dataset \cite{Cordts2016Cityscapes}.

\figE

\tabD

We performed a joint viewpoint-change prediction using each full season dataset as the map images, to compare
the proposed 
method to other comparing methods
including AD, PC, FD+AD, FD+PC, and FD+AD+PC. Table \ref{tab:D} lists the test results.
The combinations of 
templates J+B+BING+YOLO
in Fig. \ref{fig:C}
are used as 
the default setting.
Table \ref{tab:D} summarizes the result.
In the table,
``solo-leader" and ``co-leader"
represent
the ratio of query images
in which
the method outperforms the other methods
in terms of the five-grade evaluation (i.e., 
Top-5, 10, 15, 20, or otherwise).
It is evident 
that the proposed FD method combined with AD and PC (FD+AD+PC) frequently outperformed the other combinations of methods. Importantly, the FD method 
outperforms the AD and PC methods.

\section{Conclusions and Future Works}

In this paper, 
we proposed a novel method for fault diagnosis-based change detection based on 2D on-board imagery in a 3D
real-world environment. The method is substantially different from existing methods and uses a BoW image representation to
reorganize BoW flexibly into varying scales. It is shown experimentally that the proposed method boosts the accuracy of the
image change detection, 
while suppressing the map maintenance cost. Furthermore, the proposed method accounts for joint
viewpoint-change prediction by introducing strong and weak BoW features with different levels of discriminability.

Future work should address a general framework of change detection from image sequence rather than a single image. Currently, in our experimental implementation, only a single-view joint viewpoint-change prediction is considered.
Future work can leverage richer information from multi-view sequential query images \cite{FehrDLS16}. Furthermore, performance gains can be expected when change detection results from different approaches (i.e., PC, AD, FD) are combined not only in late fusion (i.e., at the level of output LoC maps) but also in early fusion 
(e.g., at the level of input features) \cite{zadeh2018multi}.

\bibliographystyle{IEEEtran}
\bibliography{icra20sugimoto}

\end{document}